\documentclass[sigconf]{acmart}
\usepackage{graphicx}
\usepackage{amsmath}
\usepackage{amsthm}
\usepackage{booktabs}
\usepackage{algorithm}
\usepackage{algorithmic}
\usepackage{multirow}
\usepackage{subfigure}
\usepackage{xcolor}
\usepackage{colortbl}
\usepackage{mathtools}
\usepackage{hhline}
\usepackage{multirow} 
\usepackage{booktabs}
\usepackage{ulem}

\AtBeginDocument{%
  }

\setcopyright{acmcopyright}
\copyrightyear{2025}
\acmYear{2025}
\acmDOI{XXXXXXX.XXXXXXX}

\acmISBN{978-1-4503-XXXX-X/18/06}

\begin{document}
\begin{sloppypar}

\title{OverFlowLight: Real-Time Gridlock Prevention and Traffic Signal Optimization for Urban Intersections}
\author{Mingyuan Li}
 \authornote{Both authors contributed equally to this research.}
 \affiliation{%
   \institution{Beijing University of Posts and Telecommunications}
   \country{}
 }

 \author{Boyang Huang}
 \authornotemark[1]
 \affiliation{%
   \institution{University of Electronic Science and Technology of China}
   \country{}
 }

 \author{Tianqi Jiang}
 \affiliation{%
   \institution{South China University of Technology}
   \country{}
 }

 \author{Chenpu Li}
 \affiliation{%
   \institution{Beijing Xiaocheng Intelligent Computing}
   \country{}
 }

 \author{Chunyu Liu}
 \affiliation{%
   \institution{Beijing University of Posts and Telecommunications}
   \country{}
 }

 \author{Yang Li}
 \affiliation{%
   \institution{Beijing University of Posts and Telecommunications}
   \country{}
 }

 \author{Ruimin Li}
 \authornote{Corresponding author.}
 \affiliation{%
   \institution{Tsinghua University}
   \country{}
 }

 \author{Qiang Wu}
 \authornotemark[2]
 \affiliation{%
   \institution{Lanzhou University}
   \country{}
 }

\begin{abstract}
Queue overflow, a severe consequence of urban traffic congestion, occurs when vehicle queues exceed intersection capacity, obstructing upstream traffic and triggering cascading gridlocks. 
Prevailing traffic signal control (TSC) algorithms, primarily optimized for throughput, often fail to address overflow during peak hours, exacerbating congestion and creating safety hazards.
We propose OverFlowLight, a real-time framework designed to preemptively resolve overflow and enhance overall TSC performance.
It first introduces a mechanism to accurately detect overflow in real-time by leveraging multi-modal sensing from cameras and radars. 
Upon detection, it dynamically generates and inserts dedicated overflow phases into the signal cycle to clear the blocking queues. 
This is orchestrated by a hybrid control design that combines rapid rule-based overflow intervention with controller back ends such as reinforcement learning (RL) for longer-horizon efficiency.
We conducted extensive real-world deployments of OverFlowLight across 43 intersections in three major cities. The framework demonstrates seamless integration with existing RL-based TSC agents, highlighting its modularity and practical applicability.
Empirical results show that OverFlowLight reduces overflow incidents by 60.4\% and increases network throughput by 18.2\% compared to deployed baselines. Furthermore, it substantially diminishes the need for manual intervention common with expert-tuned signal plans.
This work presents the first practical, scalable, and data-driven framework for actively preventing traffic gridlock, offering a crucial component for building resilient and efficient urban transportation systems.
Our demonstration videos, codes and datasets are available at the anonymous URL\footnote{\url{https://anonymous.4open.science/r/OverFlowLight-FBF9}}.

\end{abstract}




\maketitle

\begin{figure}
    \centering
    \includegraphics[width=0.95\linewidth]{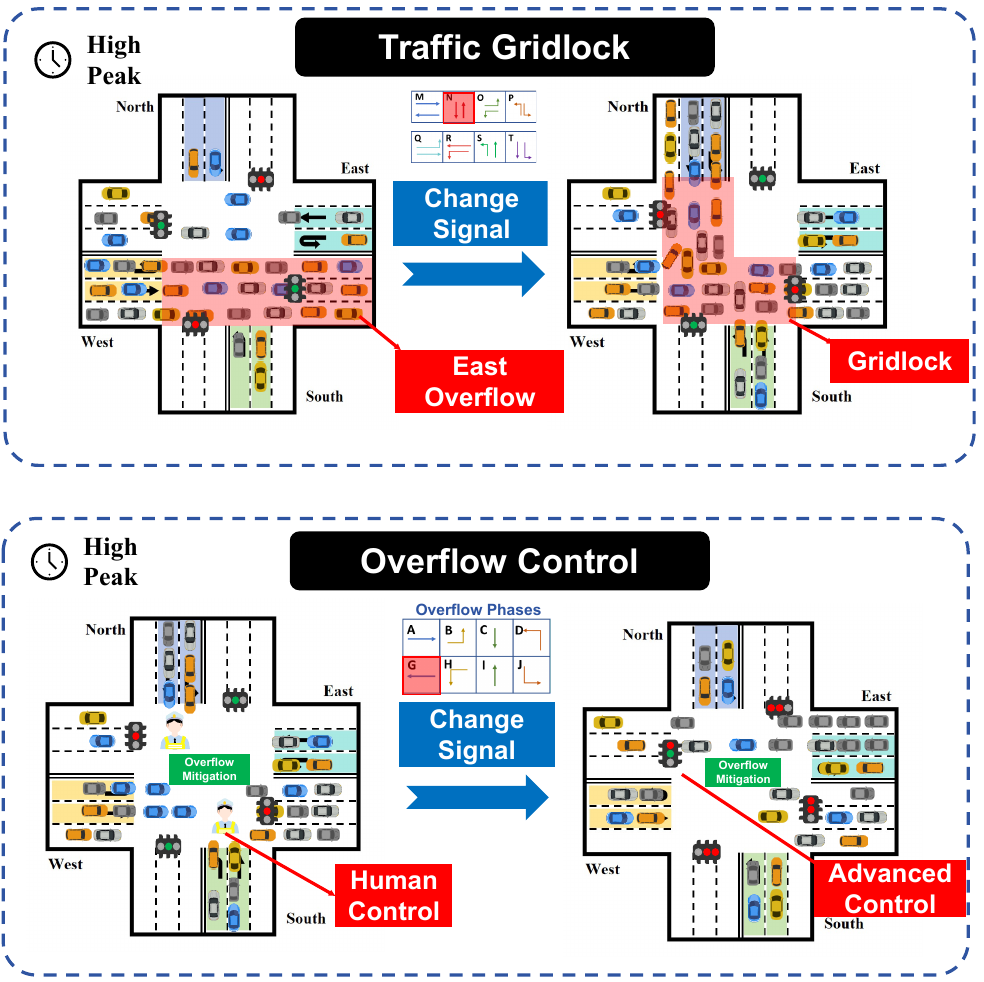}
    \caption{During peak hours, surging traffic demand can cause overflow and subsequent gridlock when signals switch. Traditional manual control is inefficient, whereas our method reduces the need for manual intervention and improves traffic efficiency.}
    \label{figure_intro}
\end{figure}

\section{Introduction}
\label{Sec:Intro}

\textbf{Motivations.} Conventional Traffic Signal Control (TSC) algorithms predominantly emphasize operational efficiency metrics under normal traffic conditions~\cite{wei2018intellilight}. 
Typical approaches include pressure maximization strategies, where the Max Pressure framework~\cite{varaiya2013max} minimizes vehicle travel times through intersection pressure optimization, and Deep Reinforcement Learning (DRL) techniques~\cite{VanDerPol2016} that employ queue length, delay time, and stopping frequency as optimization objectives~\cite{wei2018intellilight,wei2019colight}. However, these methodologies exhibit critical limitations in addressing the urban traffic phenomenon of intersection overflow and its progression to gridlock~\cite{haight1959overflow}.
As illustrated in Figure~\ref{figure_intro}, real-world deployment reveals that conventional TSC systems frequently fail under asymmetric traffic patterns, particularly during peak hours with directional flow imbalances or when intersections exhibit lane configuration mismatches (i.e., more entry than exit lanes)~\cite{akccelik1994overflow}.
Existing traffic signal control methods cannot explicitly resolve \textit{intersection overflow}, a critical failure mode in which insufficient exit-lane capacity causes vehicle queues to propagate backward into the intersection area~\cite{haight1959overflow}. 
Such overflow conditions create cascading failure risks, potentially escalating into \textit{gridlock} scenarios~\cite{kerner2011physics} where mutually obstructing vehicle clusters create a positive feedback loop, leading to the complete paralysis of intersection functionality.

\smallskip

\noindent\textbf{Challenges.} Existing TSC methodologies exhibit three critical deficiencies in addressing intersection overflow. First, existing traffic emulators fail to accurately replicate overflow conditions, and conventional TSC methods prove inadequate in addressing the gridlock issues arising from such overflow scenarios, while pressure-based optimization approaches, such as the MaxPressure algorithm~\cite{liu2022novel,mercader2020max} demonstrate partial effectiveness under moderate traffic conditions, they fundamentally fail during peak hour congestion characterized by directional flow imbalances and enter/exit lane configuration mismatches. Second, current predictive frameworks suffer from dual technical constraints: (1) reliance on imperfect traffic forecasting models with inherent estimation errors~\cite{Viti2004c,zhao2024}, and (2) insufficient temporal resolution (>30s latency) for real-time overflow prevention. Third, the absence of automated control mechanisms forces dependency on manual police intervention, this is an unsustainable solution that inefficiently allocates law enforcement resources while introducing human error risks.

These limitations collectively highlight a significant research-practice gap. Despite the severe consequences of traffic gridlock, which range from network-wide congestion cascades to heightened accident risks, existing TSC systems continue to depend on manually tailored expertise and lack a unified framework to effectively address gridlock issues stemming from overflow.

\smallskip
\noindent\textbf{Technical Overview.} OverFlowLight addresses this gap with a concise three-stage pipeline, illustrated in Figure~\ref{fig:framework}. First, the \textbf{overflow detection} module fuses radar- and camera-derived lane-level observations to identify emerging overflow directions in real time. Second, the \textbf{overflow phase construction} module maps each detected overflow direction to a feasible set of overflow-clearing phases through the overflow phase map (OPM). Third, the \textbf{OPM-constrained phase selection} module applies this safe phase set to either a traditional controller or an RL-based controller, enabling fast local intervention while preserving compatibility with existing TSC back ends. This overview gives readers the high-level logic of the system before the detailed problem formulation and method description.

\noindent\textbf{Contributions.} 
In response to the challenges enumerated above, this work presents the first real-world deployed traffic management system that systematically resolves intersection overflow through three fundamental contributions:

\begin{itemize}
    \item We formulate overflow as an actionable lane-level control problem and design an overflow-control framework that addresses gridlock during peak hours and at intersections with unbalanced lane conditions.

    \item We introduce a practical overflow detection and phase-selection interface centered on the overflow phase map (OPM), which integrates with both traditional TSC methods and RL-based controllers to mitigate gridlock caused by overflow.

     \item We conduct the first large-scale real-world validation of an overflow-preemption framework, deploying OverFlowLight across 43 intersections in three cities over 1.5 years. Empirical results demonstrate a \textbf{60.4\% reduction} in overflow incidents and an \textbf{18.2\% increase} in network throughput compared to deployed baselines, validating its effectiveness and scalability.
\end{itemize}

\section{Related Works}
\subsection{Traffic Signal Control} 

In urban traffic control, initial strategies relied on static signal timings \cite{miller1963settings}, adjusting plans for different periods like peak and off-peak hours, weekdays, weekends, and seasons. Later, heuristic methods such as Longest-Queue-First (LQF) were introduced, which switched traffic light phases based on traffic demand at intersections \cite{Papageorgiou2003}. However, these traditional approaches struggle with complex traffic scenarios and are less effective when demand varies significantly.

To overcome the shortcomings of traditional methods, Reinforcement Learning (RL) was introduced into the field of Traffic Signal Control (TSC). Methods like FRAP \cite{zheng2019learning}, IntelliLight \cite{wei2018intellilight}, Presslight \cite{wei2019presslight}, and Colight \cite{wei2019colight} leverage RL to optimize traffic signal phases, either through phase competition principles or by facilitating communication between intersections. Toward A Thousand Lights \cite{chen2020toward} leverages agent based multi intersection parameter sharing for fast convergence. AttendLight \cite{NIPS-attention} introduces a self-attention mechanism to learn phase correlations without relying on human expertise. DynamicLight \cite{zhang2022dynamiclight} adopts a two-stage framework, determining the phase with RL and then selecting the phase duration from predefined values. These RL-based methods have shown significant progress in enhancing traffic management intelligence to improve efficiency \cite{zhang2024multi,liu2023gplight}. However, they often require extensive online data and computational resources, and face challenges related to safety and stability during real-world deployment \cite{park2024traffic,ma2023traffic}.

Offline RL offers a promising alternative by learning from historical data, avoiding the need for real-time interaction with the environment. TransformerLight~\cite{wu2023transformerlight} utilizes offline RL for TSC, demonstrating the potential of this approach to improve traffic management efficiency without online experimentation. Meng et al. \cite{meng2023offline} further advance the field with an offline pre-trained Multi-Agent Decision Transformer (MADT), which enhances sample efficiency and generalization in online tasks through offline dataset pre-training. This pre-training strategy presents a novel solution for addressing complex decision-making problems in TSC \cite{han2023mitigating}, particularly in scenarios where online experimentation is costly or poses safety risks. However, most TSC methods focus primarily on improving efficiency, without addressing safety concerns.

\subsection{Traffic Overflow Prediction}
Overflow occurs when traffic intensity approaches the capacity of an intersection. In such scenarios, random fluctuations in vehicle arrival rates could result in cycle failures, where queues fail to clear within a cycle and progressively grow, ultimately forming overflow queues. Traditional queuing models have been used to describe and predict these overflow queues, but they have limitations, particularly when traffic flow is near saturation, resulting in inaccurate delay estimations \cite{Viti2004c}.

The max-pressure~\cite{varaiya2013max} method selects TSP based on the difference in vehicle counts between upstream and downstream lanes. While this approach could alleviate overflow under certain conditions, it proves ineffective in scenarios with imbalanced lane capacities between upstream and downstream or when the traffic demand in the current phase is excessively high.

To tackle these challenges, advanced queuing models have been introduced. For example,  \cite{Viti2004c} created a heuristic model that integrates initial linear trends with nonlinear elements by employing data generated through the Markov chain process. This model enhances the precision of overflow queue length calculations and delay predictions by accounting for the queue's temporal evolution and the influence of standard deviation on average queue behavior.

With the advancement of connected vehicle technology, Connected Vehicle (CV) data has emerged as a novel approach for overflow prevention. CV data, which encompasses detailed information like vehicle position, speed, and travel trajectory, could mirror the real-time state of traffic flow \cite{Li2019, Wu2020}. Certain studies have leveraged CV data to estimate traffic states, including the analysis of vehicle trajectory data to determine traffic flow, speed, and density parameters \cite{Zheng2017, Tang2020, Zhang2022}. Zhao etc. \cite{zhao2024} utilizes partial CV data, employing an adaptive Kalman filtering model in conjunction with real-time CV data to forecast the remaining space at intersection exits and vehicle arrival-departure patterns. Subsequently, it reallocates signal control using Model Predictive Control (MPC) to effectively avert overflow.

While existing methods of overflow prediction may be effective in some situations, their applicability and accuracy may be limited in situations such as infrequent congestion, limited data availability, and difficulties in obtaining real-time data. Therefore, further research is needed to develop more flexible, accurate, and practical measures for overflow prediction and prevention.

\subsection{Traffic Overflow Control}
Various solutions have been proposed in the field of traffic management for the intersection overflow problem. Early methods mainly focus on signal timing optimization, such as the lane-based optimization method proposed by \cite{wong2003lane}, which manages traffic flow by adjusting the green light time and cycle length to reduce the accumulation of vehicles while waiting. The advantage of this method is that it can significantly improve the efficiency of intersections, especially during periods of high traffic demand. However, the disadvantage of this approach is that it may ignore the actual queuing of vehicles, resulting in overflow still occurring in some cases.

Subsequently, researchers began to focus on lane marking optimization. \cite{yu2017robust} proposed a robust optimal lane assignment method to minimize the total delay at intersections by optimizing lane markings. The advantage of this method is that it can control the flow of vehicles more finely and reduce unnecessary turning and lane changing behaviors, thus reducing the risk of overflow. Nevertheless, this approach may require more complex traffic control systems and higher maintenance costs.

Most of the existing approaches are rooted in traffic engineering optimizations, relying heavily on on-site inspections by professional traffic engineers for individual intersections. These methods lack a universal, flexible signal control framework capable of effectively mitigating overflow scenarios  traffic conditions.
\section{Preliminary}

\subsection{Traffic Signal Control}
\label{defi:3.1}
\begin{figure}[ht]
        \centering
        \includegraphics[width=1\linewidth]{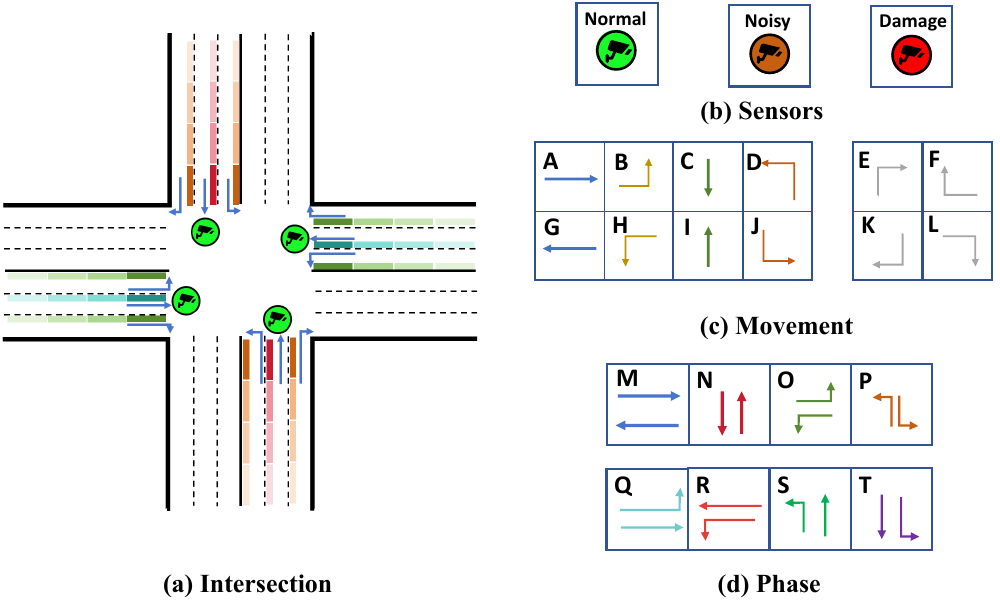}
    \caption{Definition of the TSC. Sensors are used to acquire vehicle information, which is fed into the TSC algorithm to output the appropriate phase, as shown in Figure 1(d).}
    \label{fig:definition}
\end{figure}

\begin{figure*}
    \centering
    \includegraphics[width=0.95\linewidth]{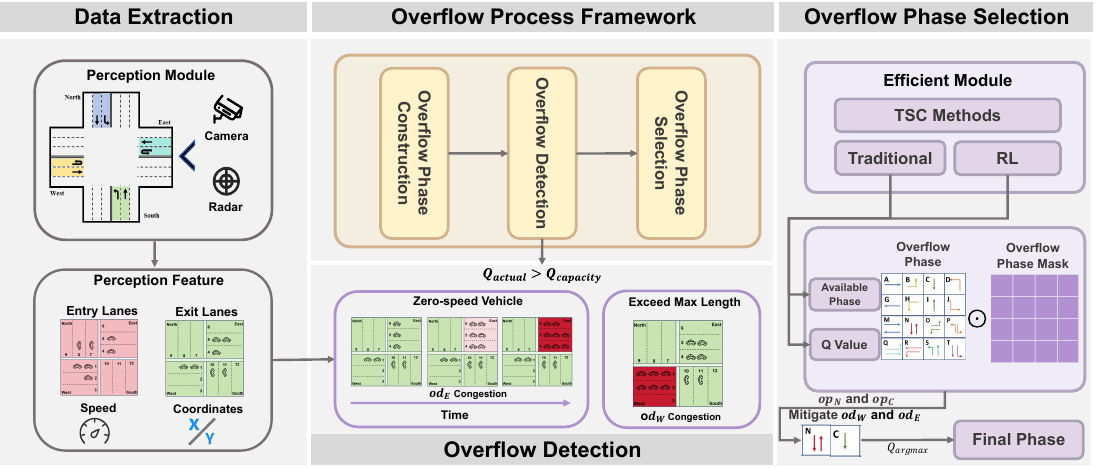}
    \caption{Overview of OverFlowLight. The framework consists of three stages.
    (1) Overflow Phase Construction: Constructs an overflow-phase map (OPM) by mapping each detected overflow direction to adaptive overflow phases. 
    (2) Overflow Detection: Extracts traffic data from cameras and radars, applying congestion metrics to identify overflow directions. 
    (3) Overflow Phase Selection: Applies a phase mask to select the optimal phase, adopting either traditional TSC or RL method.}
    \label{fig:framework}
\end{figure*}

As shown in Figure~\ref{fig:definition}. the traffic network is composed of multiple intersections and lanes.

\noindent\textbf{Intersection.}
A place where different vehicle paths cross in the traffic network. It gathers and disperses traffic from various directions. Multiple intersections combine to make up the whole traffic network, and its layout affects traffic flow. The state representation of each intersection is based on individual lanes, including metrics such as the total number of vehicles per lane ($x(l), l \in \mathcal{L}^{in}_i$), segmentation of each lane defined as $x_i(l)$ with segments spaced $m$ meters apart, and the queue length, which measures the number of waiting vehicles ($q(l), l \in \mathcal{L}^{in}_i$).

\noindent\textbf{Traffic Movement.}
This refers to vehicles crossing an intersection in a specified direction (left, straight, or right), and an intersection typically supports twelve different traffic movements, which is critical for developing effective signal phasing.

\noindent\textbf{Traffic Signal Phasing (TSP).} The heart of TSC and defines the vehicle movements that are permitted at an intersection during a specific time interval. Each TSP is associated with a signal indication that directs vehicles whether to proceed, yield, or stop. The number of phases in the network can vary, usually from two to four, and these phases are designed to manage traffic flow by sequentially assigning rights-of-way. Intersections use 2, 4, or $k$ base phases, denoted as $p=1,2,...,k$. Figure~\ref{fig:definition}(d) shows 8 phases.

\noindent\textbf{The State Representation.} It provides a detailed snapshot of traffic conditions. This representation includes metrics such as the total number of vehicles in each lane, queue lengths, and other metrics that quantify the number of vehicles waiting to move forward. Understanding these metrics allows for the development of complex TSC strategies that can adapt to real-time traffic conditions, thereby improving the efficiency and safety of the entire network.

\noindent\textbf{Traffic Perception.} Each intersection has four directional sensors (e.g., cameras, radars) monitoring vehicles in three lanes, with different security states indicated by colors: orange for noise attacks, red for sensor damage, and green for normal operation, as shown in Figure~\ref{fig:definition}(b).

\subsection{Traffic Overflow}
The overflow direction refers to the direction of vehicle behavior when the traffic flow is in a specific state. 

\noindent\textbf{Exit Overflow Direction.}
Common traffic intersections include the north exit $od_N$, south exit $od_S$, east exit $od_E$, and west exit $od_W$. Vehicles enter the intersection from these directions and participate in the traffic flow. The exit directions represent the paths vehicles take when leaving the intersection area. After turning or going straight at the intersection, vehicles exit and continue on their way. When the queue length exceeds the capacity of the exit lane, we define these exits as overflow directions, denoted as $OD=(od_N, od_i,..., od_W), i=N,...,W$.

\noindent\textbf{Overflow Direction Judgment.}
The determination of whether an overflow occurs hinges on two crucial factors: traffic volume and vehicle queue length. Regarding traffic volume, when \(Q_{actual}>Q_{capacity}\), it implies that the influx of vehicles into this exit direction surpasses its normal handling capacity, a significant contributor to potential overflow. Simultaneously, the vehicle queue length is a vital indicator. If \(L_{queue}>L_{threshold}\), it indicates that the vehicle accumulation at the exit has exceeded the normal range. Only when both these criteria are met, along with additional factors such as unreasonable signal timing (\(t_{green}\) being too short), traffic accidents, or other interferences preventing normal vehicle evacuation at the exit, and vehicles start spreading around the exit, can it be concluded that an overflow has happened, which is then defined as the Overflow Direction.

\noindent\textbf{Overflow Phase.} The overflow phase is defined as a specific phase in a traffic TSC that mitigates the overflow direction. Each individual movement, as well as the combination of non-conflicting movements, can be considered as an overflow phase. \textbf{The combination of all movements Figure~\ref{fig:definition}(c) and phases Figure~\ref{fig:definition}(d) constitutes the list of overflow phases $op = 1,2,3,4,...,m$}.

The reason for considering individual movements as overflow phases is that overflow may occur several seconds before the current phase ends. For example, at overflow direction $od_E$, the current phase might be phase M Figure~\ref{fig:definition}(d), but if there are still many vehicles in movement G Figure~\ref{fig:definition}(c), it could result in the inability to clear the vehicles in that lane before the phase ends. This situation could cause the overflow to propagate to the next intersection in movement G, leading to congestion there as well. When the vehicles in movement G are not cleared in time due to insufficient green time in the current phase, the overflow extends to the following intersection, where the same vehicles might still be waiting. This can cause a ripple effect, worsening congestion downstream.

As shown in Figure~\ref{figure_intro}. If direction $od_E$ experiences heavy traffic, an overflow phase might be triggered to movement G and allow more time for mitigating overflow diretion $od_E$. Once the overflow is resolved, the algorithm will switch to the phase $p$ that can improve efficiency.

\section{Methods}
\label{Sec:Methods} 
In this section, we present \textbf{OverFlowLight}, a real-time overflow control framework designed to mitigate vehicle spillovers and optimize Traffic Signal Control (TSC) at urban intersections. \textbf{OverFlowLight} comprises three core modules: Overflow Phase Construction, Overflow Detection, and Overflow Phase Selection.

\subsection{Overflow Phase Construction}
To construct overflow phases, we design an algorithm that adjusts signal phases based on traffic overflow conditions. The algorithm starts with the \textbf{base phases $p$} (Definition~\ref{defi:3.1}), which define the fundamental traffic signal plans under normal conditions and serve as a reference for adjustments.

The algorithm generates all available \textbf{overflow phases $op$}, representing alternative movement and phases to address overflow situations. These phases adapt dynamically to changing traffic patterns and improve intersection efficiency during congestion.

In addition, the algorithm incorporates \textbf{overflow directions} (e.g., $od_{E}$, $od_{S}$), which indicate where vehicle accumulation is most significant, guiding the selection of effective overflow phases.

The algorithm operates as follows:
\begin{algorithm}[t]
\caption{Overflow Phase Map Construction}
\label{alg:overflow_phase_core}
\begin{algorithmic}[1]
    \REQUIRE 
   \( D \) List of all directions \( d \)\\
\ENSURE Overflow phase map $OPM$
\STATE Initial map $OPM$ 
    \FOR{each \( d \) in \( D \)}
    \STATE For each direction $d$, initial list $OPM[d]$
        \FOR{each \( op \) in \( OP \)}
        \IF{$op$ could mitigate $d$}
                \STATE $OPM[d]$ add $op$
            \ENDIF
        \ENDFOR
    \ENDFOR
    \STATE Return \( OPM \)
\end{algorithmic}
\end{algorithm}
The algorithm initializes a list of overflow phases for each overflow direction $d$, and iterates through the overflow phases to identify an available overflow phase $op$ that can mitigate the overflow in the specified direction. For example, overflow phase $op_G$ Figure~\ref{fig:definition}(c) could mitigate overflow direction $od_E$. From this, we obtain a mapping from overflow direction to overflow phase:
\[
\text{overflow phase map $OPM$}: OPM[d] \mapsto op
\]

This map provides guidance for subsequent overflow selection.

\subsection{Overflow Detection}

Overflow detection is divided into three steps: Data Extraction, Applying Overflow Rules, and Get Overflow Mask.

\subsubsection{Data Extraction: }
Surveillance cameras and radars are installed at each directions of intersections to monitor traffic conditions in real-time. We detect data within an effective range $ER$ of each direction. The \textbf{Perception Module} collects data from entry and exit lanes, including the number of queuing vehicles, their speed, and waiting time to assess if the direction is overflow states.

The \textbf{Perception Feature} further processes this data, providing:
\begin{itemize} \item \textbf{Vehicle Speed:} Real-time speed of each vehicle in each lane to better analyze waiting vehicles.
\item \textbf{Vehicle Coordinates:} Positions of vehicles based on a coordinate system centered at the intersection, allowing precise vehicles within entry and exit lanes.
\end{itemize}

\subsubsection{Apply Overflow Rules:}
The collected monitoring data is then compared against the pre-defined overflow conditions. We have established two key overflow conditions:
\begin{enumerate}
    \item \textbf{Waiting  Condition}: The number of waiting vehicles and the average waiting time on most lanes of a particular road direction exceed the specified thresholds.
    \item \textbf{Queuing Congestion Condition}: The number of queuing vehicles (low-speed vehicles) on a specific road direction surpasses the set threshold.
\end{enumerate}

These two conditions, by monitoring and responding to traffic conditions in real time, enable early congestion warnings, optimize traffic flow, and prevent congestion from spreading, thereby effectively alleviating traffic congestion.
\subsubsection{Get Overflow Mask:}
If either of Apply Overflow Rules conditions is satisfied, the overflow direction is then add to the list of available overflow directions 
 to construct overflow phase mask $M \in \{0, 1\}^{1 \times m}$, where $m$ is the count of overflow phase.

The detection function iterates through each exit road and its corresponding lanes, applying Algorithm~\ref{alg:overflow_detection} to generate the overflow mask. The Algorithm~\ref{alg:overflow_detection} iterates through vehicles in each overflow direction $OD$ from lines 3 to 11, applies overflow rules to determine the available overflow phase indices based on the overflow phase map $OPM$, and subsequently marks the overflow mask corresponding to these indices as 1. If no overflow is detected, the overflow mask for the base phases is set to 1. Finally, the algorithm outputs the overflow phase mask $M$ based on overflow flag $OF$.
\begin{algorithm}[h]
\caption{Get Overflow Phase Mask}
\label{alg:overflow_detection}
\begin{algorithmic}[1]
    \REQUIRE 
        \( OD \) Set of overflow directions \\
        \( \alpha \): Zero-speed vehicle waiting time threshold \\
        \( \beta \): Max exit vehicle count  \\
        \( EP \): Effective range to detect overflow directions
    \ENSURE 
        \( M \): Overflow phase mask and overflow flag $OF$

    \STATE Initial overflow flag $OF \leftarrow False$  
    \STATE Initial overflow phase mask $M \leftarrow 0$
    \FOR{each \( od \) in \( OD \)}
        \STATE   Calculated max queued vehicles $V_q$ within $EP$ 
        \STATE Calculated max vehicle waiting time $V_z$ within $EP$
        \IF{ \( V_q \geq \beta \) \OR \( V_z \geq \alpha \) }
          \STATE Get list of overflow phases index $OPI$ from $OPM[od]$
          \STATE Set overflow phase mask $M_{OPI} \leftarrow 1$
          \STATE Set overflow flag $OF \leftarrow True$
        \ENDIF
    \ENDFOR
    \IF{$OF$ is $False$}
    \STATE Get base phases index $BPI$ 
    \STATE Set overflow phase mask $M_{BPI} \leftarrow 1$
    \ENDIF
    \RETURN \( M \textbf{ and } OF\)
\end{algorithmic}
\end{algorithm}
\subsection{Overflow Phase Selection}
The final module of \textbf{OverFlowLight} involves selecting an appropriate overflow phase based on the generated overflow phase mask $M$. This module is pivotal as it determines the traffic signal adjustments required to mitigate overflow and prevent gridlock. \textbf{OverFlowLight} provides a unified interface for integrating different control strategies, including both traditional TSC methods and Reinforcement Learning (RL)-based approaches. 

\subsubsection{Traditional TSC Approach}
In traditional TSC methods, available overflow phases are selected based on predefined criteria prioritizing traffic flow efficiency. For instance, the Advanced-MaxPressure method evaluates the pressure on each lane and selects phases that alleviate the most significant congestion.

\subsubsection{RL-based Overflow Control Algorithm}
Building upon the \textbf{OverFlowLight} framework, we introduce a novel RL-based overflow control algorithm that optimizes phase selection dynamically. The RL algorithm is designed with the following components:

\begin{itemize}
\item[$\bullet$]\textbf{State.}
    The state is the number of waiting vehicles and waiting vehicles within $ER$. 
    \item[$\bullet$]\textbf{Action.} The action $A$ is all of the overflow phase $OP$.
    \item[$\bullet$]\textbf{Reward.} Negative queue length $r_i=\sum{-q(l)}, l \in \mathcal{L}^{in}_i  $ is denoted as the reward.
\end{itemize}
\begin{algorithm}[ht]
\caption{Overflow Phase Selection}
\label{alg:overflow_selection}
\begin{algorithmic}[1]
    \REQUIRE 
        Overflow phase map \( OPM \)
    \ENSURE Final phase \( FP \)
    \IF{ Traditional method}
    \STATE Get available overflow phases as action $\hat{A} \leftarrow OP \odot OPM$
    \STATE Consider efficiency and $\hat{A}$ to get final phase $FP$ 
    \ENDIF
        
    \IF{ RL method}
    \STATE Get available Q value $Q_{op} \leftarrow Q_{op} \odot OPM$
    \STATE Get final phase $FP \leftarrow \arg\max_{a \in A} Q(s, a)$ 
    \ENDIF
    
    \RETURN \( FP \)
\end{algorithmic}
\end{algorithm}
The RL agent learns an optimal policy to minimize congestion and prevent gridlock by selecting overflow phases. Algorithm~\ref{alg:overflow_selection} outlines this process, where traditional TSC methods (lines 2 and 4) obtain available overflow phases and their Q-values, effectively mitigating intersection overflow during peak hours using either traditional or RL-based approaches.


The efficacy of this RL-based approach is underpinned by its convergence properties, which are rooted in the Bellman equations~\cite{watkins1992q}:
$$Q^\pi(s_t, u_t) = \mathbb{E}_{r_t, s_{t+1} \sim E} [r_t + \gamma \mathbb{E}_{u_{t+1} \sim \pi} [Q^\pi(s_{t+1}, u_{t+1})]]$$

Our RL agent aims to find the optimal action-value function $Q^*(s,a) = \max_{\pi} Q^\pi(s,a)$. This optimal function satisfies the Bellman optimality equation, where the inner expectation over actions under policy $\pi$ (i.e., $\mathbb{E}_{u_{t+1} \sim \pi}$) is replaced by a maximization over all possible next actions. Expressed in terms of our problem's notation:
$$Q^*(s, a) = \mathbb{E}_{s'} [r_i + \gamma \max_{a'} Q^*(s', a') | s, a]$$
The convergence of our Q-learning algorithm to this $Q^*(s,a)$ is guaranteed under standard conditions:
\begin{enumerate}
    \item The action space $A$ and the state space $S$ are discrete and finite. Specifically, the state space is rendered finite by categorizing vehicle counts based on predefined thresholds.
    \item All relevant state-action pairs $(s,a)$, particularly those permitted by the $OPM$ as indicated in Algorithm~\ref{alg:overflow_selection}, are sufficiently explored.
    \item Appropriate values for the learning rate $\alpha$ and discount factor $\gamma$ are applied.
\end{enumerate}

The $OPM$ plays a crucial role in this learning process. By constraining the agent's actions to a contextually relevant subset $A_{OPM}(s)$, the OPM guides the algorithm to converge to $Q^*_{OPM}(s,a)$, the optimal Q-values within this OPM-constrained policy space, which inherently reflects safer and more effective operational strategies. This guidance not only upholds the convergence guarantee but also promotes more stable and potentially faster learning towards a practically sound solution.

\begin{table*}[ht]
\centering
\setlength{\extrarowheight}{0pt}
\addtolength{\extrarowheight}{\aboverulesep}
\addtolength{\extrarowheight}{\belowrulesep}
\setlength{\tabcolsep}{1mm}
\setlength{\aboverulesep}{0pt}
\setlength{\belowrulesep}{0pt}
\caption{Performance Comparison Across Peak Hours and Control Methods (For Overflow Count and Speed Variance, the smaller the better; for Speed Mean, Speed Median, and Throughput, the higher the better).}
\begin{tabular}{cccccc} 
\toprule
\multirow{2}{*}{Metric}         & \multirow{2}{*}{Dataset} & \multicolumn{2}{c}{\textbf{Morning Peak-Hour}}                                                              & \multicolumn{2}{c}{\textbf{Evening Peak-Hour}}                                                               \\ 
\hhline{~~----}
                                &                          & Traditional Fixed & {\cellcolor[rgb]{0.753,0.753,0.753}}FuzzyLight (\textbf{Improvement})                   & Traditional Fixed & {\cellcolor[rgb]{0.753,0.753,0.753}}DQN (\textbf{Improvement})                           \\ 
\hhline{------}
\multirow{2}{*}{Overflow Count} & Intersection 1           & 75                & {\cellcolor[rgb]{0.753,0.753,0.753}}27 (\textbf{+64.0\%})                               & 81                & {\cellcolor[rgb]{0.753,0.753,0.753}}36 (\textbf{+55.6\%})     \\
                                & Intersection 2           & 88                & {\cellcolor[rgb]{0.753,0.753,0.753}}42 (\textbf{+52.3\%})                      & 69                & {\cellcolor[rgb]{0.753,0.753,0.753}}16 (\textbf{+76.8\%})     \\ 
\hline
\multirow{2}{*}{Speed Mean (m/s)}     & Intersection 1           & 2.81              & {\cellcolor[rgb]{0.753,0.753,0.753}}3.56 (\textbf{+26.7\%})  & 2.76              & {\cellcolor[rgb]{0.753,0.753,0.753}}3.82 (\textbf{+38.4\%})   \\
                                & Intersection 2           & 3.54              & {\cellcolor[rgb]{0.753,0.753,0.753}}5.341 (\textbf{+50.9\%}) & 4.81              & {\cellcolor[rgb]{0.753,0.753,0.753}}6.42 (\textbf{+33.5\%})   \\ 
\hline
\multirow{2}{*}{Speed Variance (m/s)} & Intersection 1           & 14.64             & {\cellcolor[rgb]{0.753,0.753,0.753}}11.79 (\textbf{-19.5\%}) & 13.98             & {\cellcolor[rgb]{0.753,0.753,0.753}}12.33 (\textbf{-11.8\%})  \\
                                & Intersection 2           & 15.6              & {\cellcolor[rgb]{0.753,0.753,0.753}}14.14 (\textbf{-9.4\%})  & 15.4              & {\cellcolor[rgb]{0.753,0.753,0.753}}13.48 (\textbf{-12.5\%})  \\ 
\hline
\multirow{2}{*}{Speed Median (m/s)}   & Intersection 1           & 0.9               & {\cellcolor[rgb]{0.753,0.753,0.753}}2.95 (\textbf{+227.8\%}) & 2.53              & {\cellcolor[rgb]{0.753,0.753,0.753}}3.72 (\textbf{+47.0\%})   \\
                                & Intersection 2           & 2.5               & {\cellcolor[rgb]{0.753,0.753,0.753}}5.5 (\textbf{+120.0\%})  & 4.17              & {\cellcolor[rgb]{0.753,0.753,0.753}}6.28 (\textbf{+50.6\%})   \\ 
\hline
\multirow{2}{*}{Throughput}     & Intersection 1           & 2148              & {\cellcolor[rgb]{0.753,0.753,0.753}}2921 (\textbf{+36.0\%})  & 2438              & {\cellcolor[rgb]{0.753,0.753,0.753}}2452 (\textbf{+0.6\%})    \\
                                & Intersection 2           & 3146              & {\cellcolor[rgb]{0.753,0.753,0.753}}3941 (\textbf{+25.3\%})  & 2833              & {\cellcolor[rgb]{0.753,0.753,0.753}}3168 (\textbf{+11.8\%})                                  \\
\bottomrule
\end{tabular}

\label{tab:combined_results}

\end{table*}

\section{Experiments}
We deploy our overflow algorithm at 43 real-world intersections in three cities and evaluate the proposed overflow phase-selection method using FuzzyLight as an efficiency-enhancement baseline. Specifically, we conduct a comprehensive performance analysis during peak hours, when overflow events are most likely to occur. Over a 1.5-year experimental period, including rain, snow, and fog conditions, we select two critical intersections in City 2 that are particularly prone to overflow during peak hours and compare their performance against expert-designed timing plans. To improve readability, we organize this section into four parts: core quantitative performance, ablation and sensitivity analysis, representative case studies, and deployment cost, latency, and robustness. More details on the settings are provided in~\textbf{Appendix A}.

\subsection{Core Quantitative Performance}

\begin{table}[t]
\centering
\caption{Aggregate improvements across 43 deployed intersections relative to expert-timed control.}
\label{tab:deployment_summary}
\begin{tabular}{lccc}
\toprule
Metric & Peak Hours & Normal Hours & Off-Peak \\
\midrule
Throughput ($\uparrow$) & +15.8\% $\pm$ 3.2\% & +8.4\% $\pm$ 2.1\% & +1.5\% $\pm$ 0.5\% \\
Stops ($\downarrow$) & -18.7\% $\pm$ 4.5\% & -7.9\% $\pm$ 2.8\% & -2.2\% $\pm$ 0.6\% \\
\bottomrule
\end{tabular}
\end{table}

\subsection{Real-World Results} 
As shown in Table~\ref{tab:combined_results}, FuzzyLight reduces the overflow count by 64.0\% at Intersection 1 and 52.3\% at Intersection 2 compared with the traditional fixed-duration method. The DQN-based controller reduces the overflow count by 55.6\% at Intersection 1 and 76.8\% at Intersection 2 during the evening peak hour compared with the same baseline. Table~\ref{tab:deployment_summary} complements these stress-test intersections with deployment-level aggregate statistics across all 43 intersections, showing that the gains generalize beyond the two case studies and are largest during peak periods.

\subsection{Ablation and Sensitivity Analysis}
We further analyze the contribution of the major modules and the sensitivity of the overflow-detection thresholds. In particular, we examine the effect of removing OPM-based phase masking or real-time overflow detection, and we study the robustness of the key thresholds $\alpha$ and $\beta$ around their default settings. Due to space constraints, we place the detailed plots and extended comparisons in the appendix and anonymized supplementary materials, while preserving the main outcome here: OverFlowLight remains stable across reasonable threshold ranges, and performance degrades primarily when the thresholds become overly aggressive or overly conservative.

\begin{table}[t]
\centering
\caption{Overflow Switch Success (OSS) for advanced controllers with and without OverFlowLight under real-traffic replay at Intersection 1.}
\label{tab:advanced_baselines}
\begin{tabular}{lcccc}
\toprule
Method & 07-14 & 07-15 & 07-16 & 07-17 \\
\midrule
O-AdvCoLight & 94.54\% & 92.86\% & 81.48\% & 100\% \\
AdvCoLight & 13.45\% & 12.57\% & 13.46\% & 9.34\% \\
O-AdvMaxPressure & 94.54\% & 92.86\% & 81.48\% & 100\% \\
AdvMaxPressure & 10.73\% & 9.43\% & 12.46\% & 11.25\% \\
\bottomrule
\end{tabular}
\end{table}

\begin{figure*}[ht]
    \centering
    \includegraphics[width=1\linewidth,height=1.3in]{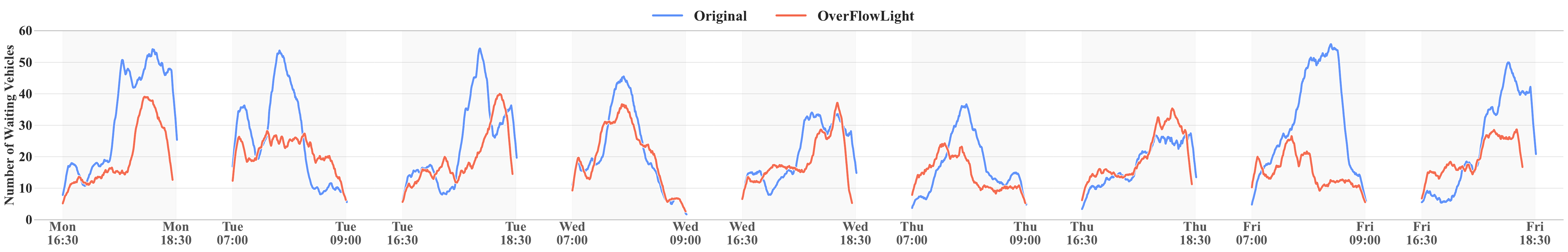}
    \caption{Peak-hour waiting-vehicle counts over one week after moving-average smoothing. The blue line represents the original method, while the red line represents our OverFlowLight algorithm.}
    \label{fig:waiting_vehicles}
\end{figure*}
\begin{figure*}[ht]
	\centering
    
        
    
\begin{minipage}{1\linewidth }
		\subfigure[Congestion]{
			\label{fig:3}
			\includegraphics[width=0.24\linewidth,height=1.2in]{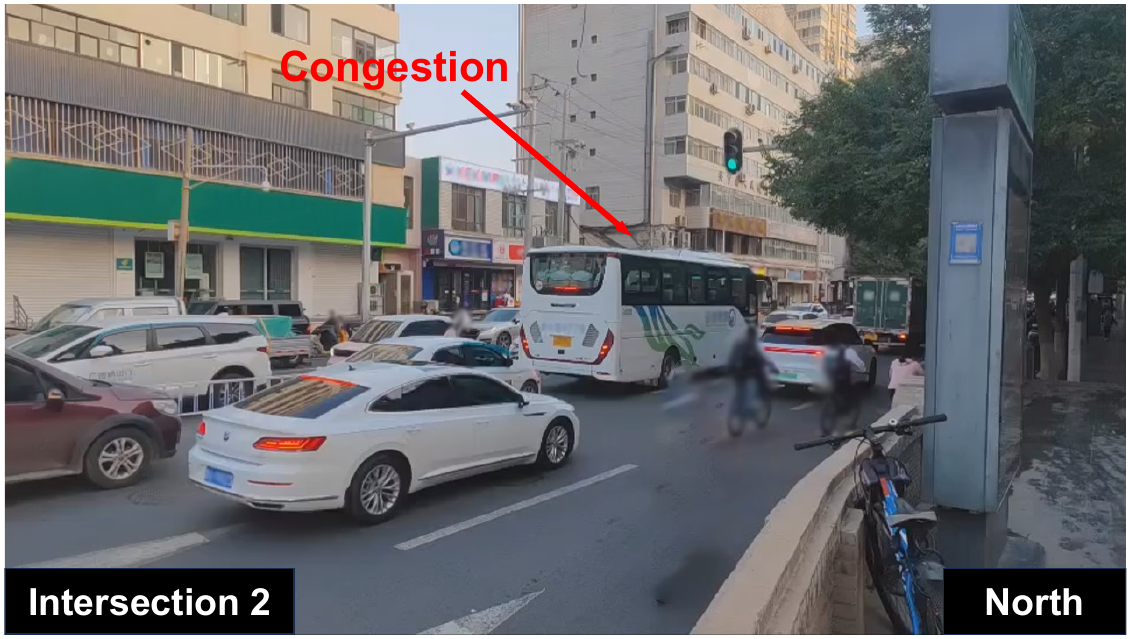}
		}\noindent
\subfigure[$d_N$ warning]{
			\label{fig:4}
			\includegraphics[width=0.24\linewidth,height=1.2in]{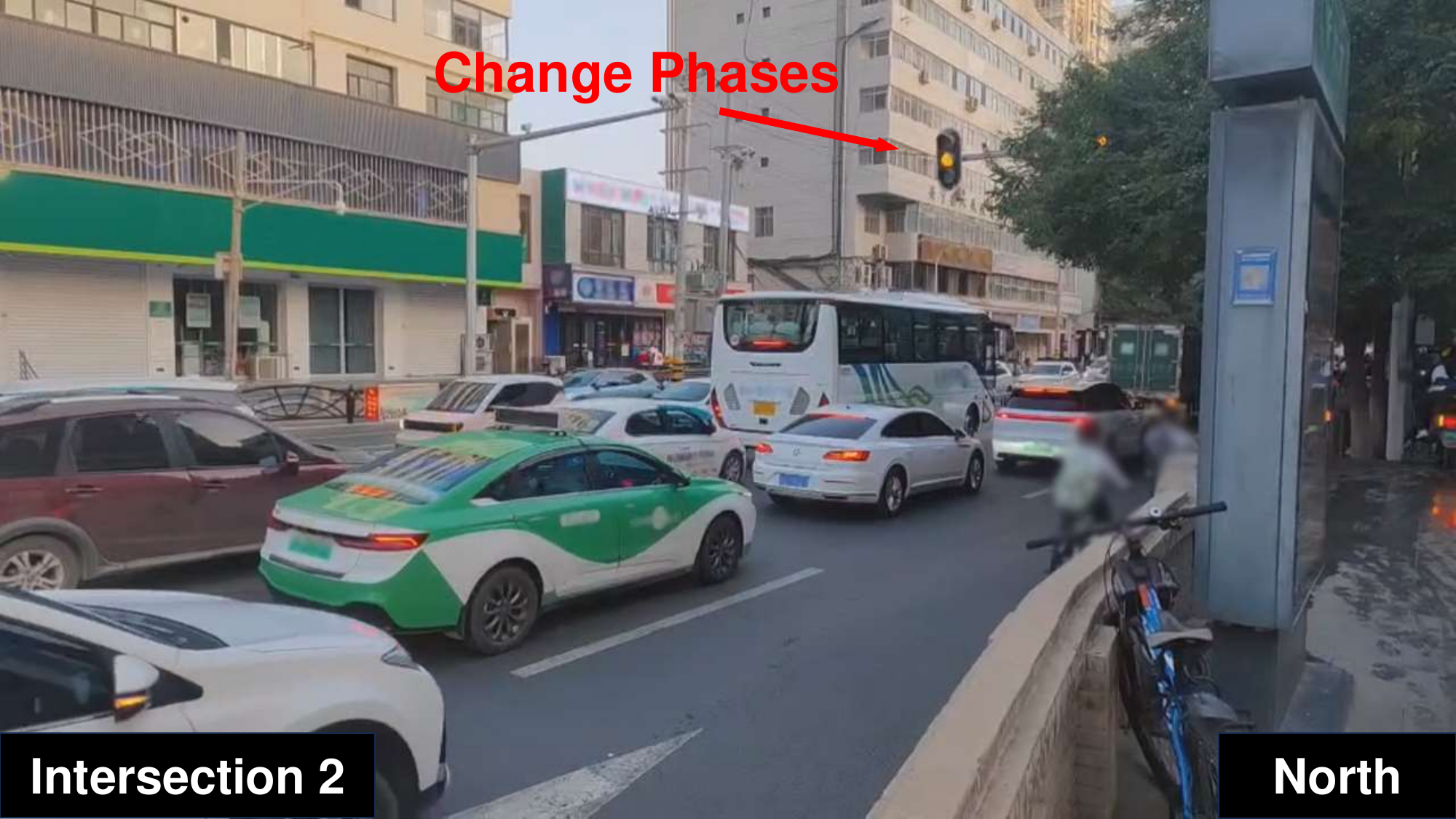}
			}\noindent
    \subfigure[Change phase to $op_I$]{
			\label{fig:3}
			\includegraphics[width=0.24\linewidth,height=1.2in]{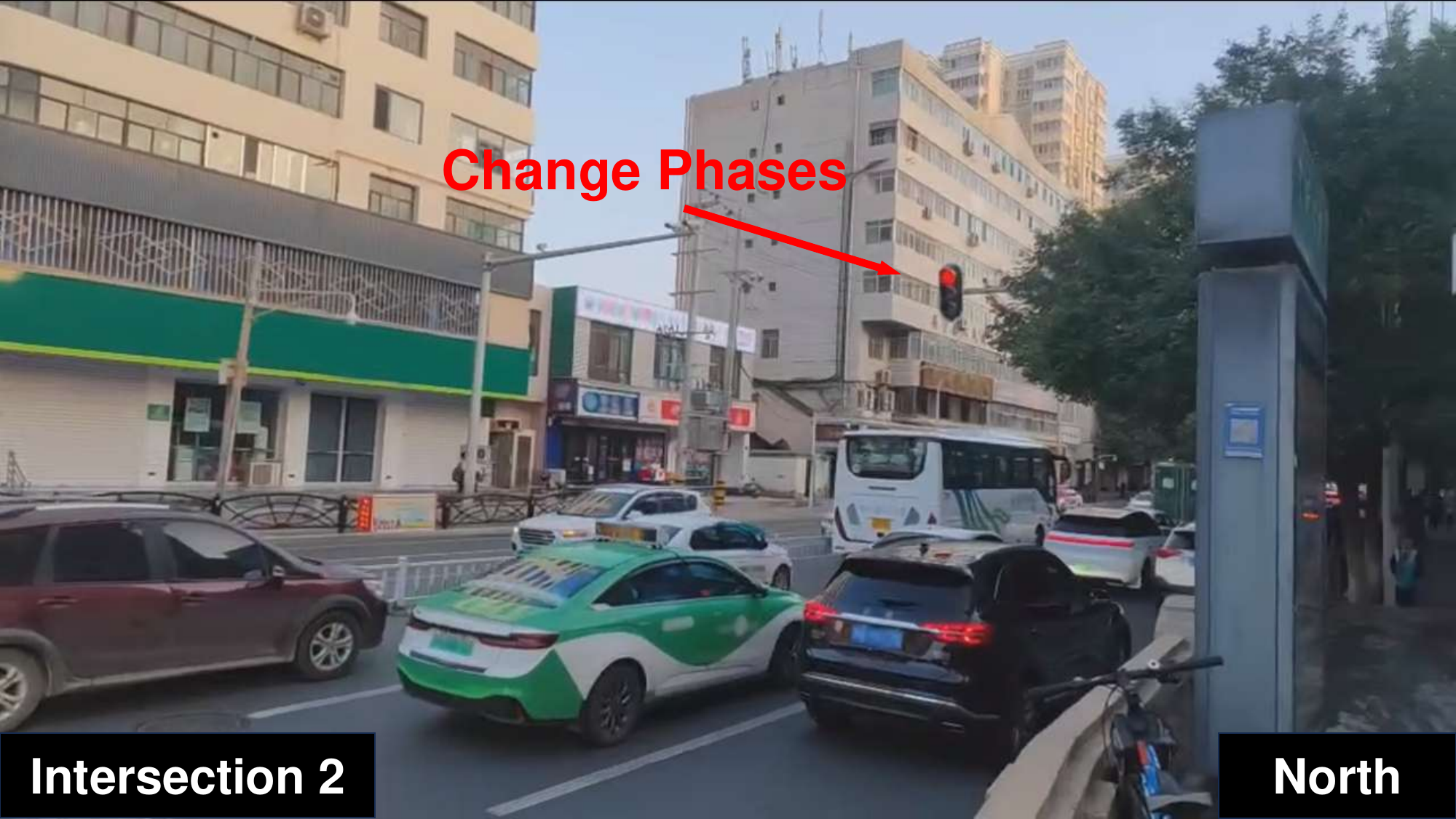}
		}\noindent
		\subfigure[Change phase to $op_I$]{
			\label{fig:4}
			\includegraphics[width=0.24\linewidth,height=1.2in]{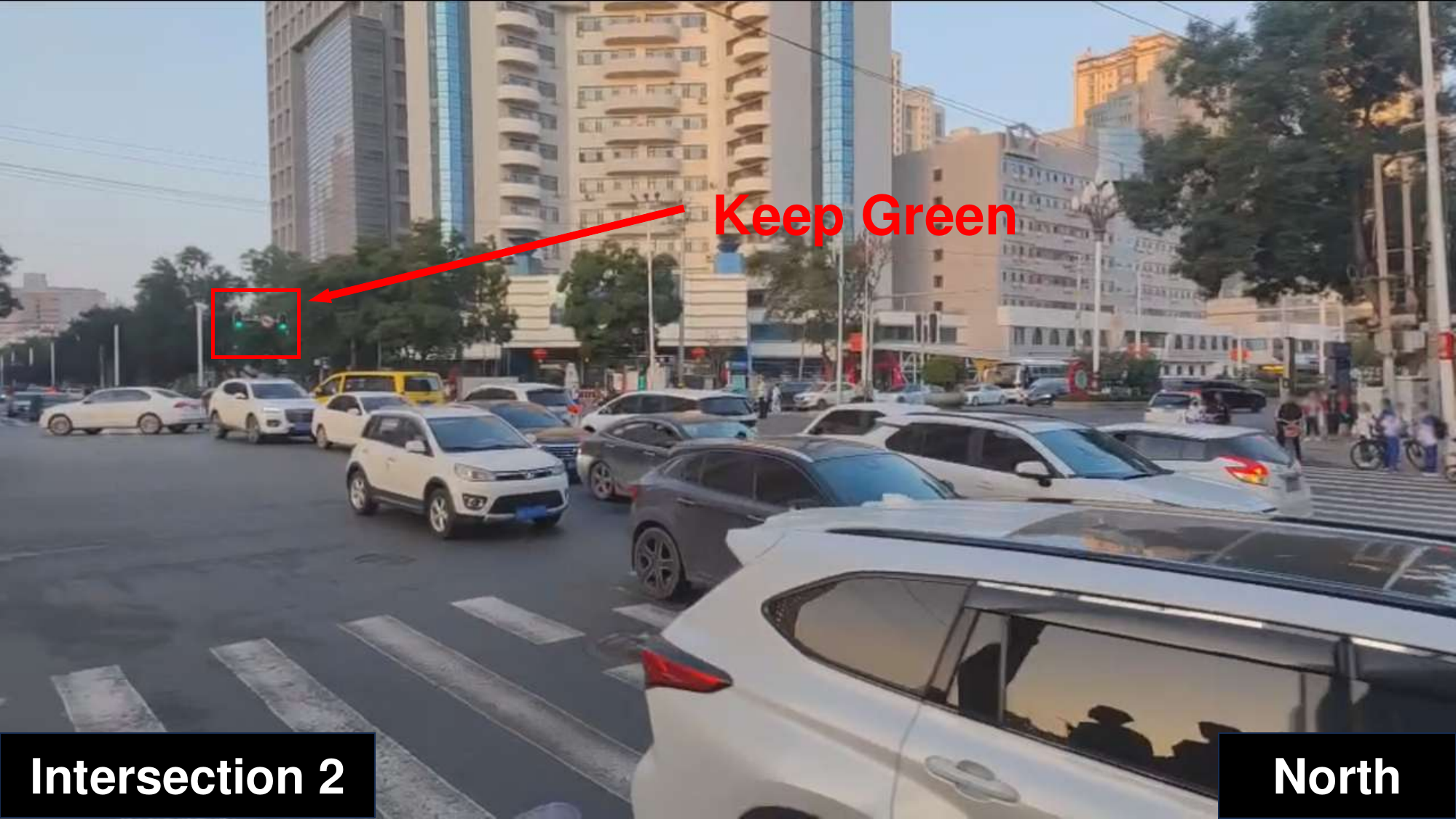}
			}\noindent
	\end{minipage}

	\caption{Real-world signal-control snapshots at Intersection 2 integrated with our proposed overflow framework.}
	
	\label{fig:intersection1}
\end{figure*}

Table~\ref{tab:advanced_baselines} provides an explicit with-versus-without comparison for advanced controllers. Integrating OverFlowLight increases overflow-switch success from roughly 10\%--13\% to 81\%--100\% across representative dates. This indicates that the main limitation of existing advanced TSC methods in these settings is not generic phase optimization, but the lack of an explicit overflow-handling mechanism.

\subsection{Representative Case Studies}
As shown in Figure~\ref{fig:waiting_vehicles}, we recorded the waiting-vehicle volume at a representative intersection after applying FuzzyLight during both morning and evening peak periods. The data reveal several patterns: the morning peak typically occurs between 7:00 AM and 8:00 AM, while the evening peak appears between 6:00 PM and 6:30 PM. Notably, the number of waiting vehicles during these peak periods is significantly lower on weekends than on weekdays.

Figure 5(a)-(b) demonstrates a case study of FuzzyLight's overflow control mechanism during peak traffic hours at Intersection 1. The results indicate that our algorithm dynamically switches to the overflow phase 
$op_N$ upon detecting overflow in the $d_w$ direction. Similarly, Figures 5(c)-(f) showcase a case study of simple DQN control at Intersection 2 during peak hours. The algorithm effectively addresses congestion and overflow in the $d_n$ direction by transitioning to the overflow phase $op_I$, thereby alleviating the overflow phenomenon.

\subsection{Deployment Cost, Latency, Robustness, and Implementation Details}
The computational pipeline achieves an average latency of 35.61 ms for the perception module and 10.56 ms for the overflow-detection module. With the full DQN-based system averaging approximately 106 ms end-to-end, it satisfies the requirements of real-time control. We also evaluate the system under long-term real-world operation, including rain, snow, fog, and other challenging conditions over 1.5 years of deployment. These results indicate that the radar-centered multimodal sensing stack remains reliable in adverse environments while preserving practical deployment cost and low-latency execution.

\begin{table}[t]
\centering
\caption{Compact deployment-cost summary per intersection.}
\label{tab:deployment_cost_main}
\scalebox{0.8}{
\begin{tabular}{lcc}
\toprule
Mode & Cost / Intersection & Practical Role \\
\midrule
Radar mode & USD 2,000--2,600 & High-reliability deployment \\
Camera-only mode & USD 400--1,000 & Low-cost scalable deployment \\
Edge compute only & USD \phantom{0}200 & 4-camera perception + RL control \\
\bottomrule
\end{tabular}
}
\end{table}

\noindent\textbf{Deployment insights.}
In practice, the hardware cost is dominated by sensing rather than control or compute. Radar provides strong robustness under nighttime, rain, and occlusion, making it appropriate for safety-critical or congestion-prone intersections. Camera-only deployment can substantially reduce hardware cost by reusing existing infrastructure, though it is more sensitive to lighting conditions. The control stack itself remains lightweight and does not require a GPU at inference time, enabling deployment on low-cost CPU/NPU edge devices. In large networks, a hybrid strategy is most practical: radar is reserved for high-risk intersections, while standard intersections use camera-only deployment. This design yields an estimated 40\%--60\% cost reduction while preserving robustness and scalability. Detailed component-wise and city-scale budget tables are provided in the appendix.

\label{Sec:Exp}
 
\section{Discussion}
\textbf{(1) What are the key innovations, and generalizability of the proposed overflow control framework compared to existing methods, and how is its superiority demonstrated?} 
Our proposed overflow control framework introduces several key innovations, such as lane-level overflow formalization, adaptive overflow phases, and OPM-constrained integration with existing RL and rule-based controllers. The algorithm demonstrates strong generalizability, having been successfully applied to multiple real-world intersections. It significantly reduces overflow occurrences, improves traffic-flow efficiency, and minimizes the need for manual intervention, outperforming traditional methods in deployment.
\\
\noindent\textbf{(2) Why not conduct more experiments on simulators, is the experimental design sufficient to support the conclusions, and are there plans for larger-scale validation and alignment analysis with expert-designed signal plans?} The reason for not conducting more experiments on simulators like SUMO or CityFlow~\cite{cityflow} is that these platforms are currently unable to accurately simulate overflow conditions. We prioritize real-world applicability and have validated the algorithm's effectiveness through extensive real-world deployments at 43 intersections across three cities. In addition, we have successfully achieved overflow control during peak hours with continuous 24/7 operation and have gained recognition from local authorities. In future work, we plan to deploy our algorithm across more intersections to further validate its scalability and effectiveness.

\noindent\textbf{(3) How are sensor data noise, overflow detection thresholds (e.g., $\alpha$ and $\beta$), and RL training processes addressed to ensure robustness and generalizability across traffic scenarios?} In the RL training process, the accuracy of perception data is crucial. To address this, we use higher-precision radar equipment to ensure accurate perception. Additionally, we employ compressed sensing from FuzzyLight to reduce noise generated during communication. The hyperparameters, such as $\alpha$ and $\beta$, were determined through extensive overflow testing, and the added sensitivity study confirms stable performance around the default settings. Regarding generalizability, the roads and lanes in our dataset are highly imbalanced across intersections. Nevertheless, the proposed overflow framework can be integrated with both traditional TSC methods and RL approaches and has been successfully deployed in real-world scenarios. This demonstrates its strong adaptability and effectiveness across diverse traffic conditions.

\noindent\textbf{(4) How does the OverFlowLight framework address multi-intersection coordination, and what are the plans for achieving network-level traffic optimization?}
OverFlowLight currently prioritizes robust single-intersection optimization to effectively manage localized overflow phenomena, while its underlying architecture is designed with network-level awareness and future scalability in mind. An established communication infrastructure enables individual OverFlowLight-equipped intersections to exchange critical traffic state information, such as exit lane vehicle speeds and capacity data from adjacent intersections, via a central server. Our empirical assessments show this data exchange occurs with a low latency of approximately 0.1 seconds, providing a foundation for future context-aware control. 

\noindent\textbf{(5) How does the OverFlowLight framework address practical deployment considerations such as hardware dependencies, cost-effectiveness, data privacy, and its chosen sensor architecture?}
The OverFlowLight framework strategically addresses practical deployment considerations by balancing advanced traffic-management capabilities with economic viability and robust operation. Recognizing hardware dependencies as a potential barrier, the system primarily employs cost-effective 80GHz millimeter-wave radar, complemented by event-based cameras, and integrates visual data anonymization techniques. This approach successfully controls per-intersection deployment costs to approximately several hundred U.S. dollars while addressing data privacy. The chosen sensor architecture leverages a multi-sensor fusion strategy integrating data from radar, event cameras, and conventional cameras to harness their complementary strengths and mitigate individual limitations. We emphasize that the VLM-based camera-only path is an optional extension rather than the default deployed stack.

\noindent\textbf{(6) How does OverFlowLight assert its academic novelty, particularly given the established use of DQN in TSC and in comparison to emerging large model approaches?}
OverFlowLight's novelty lies not in a new RL algorithm, but in its pioneering real-world system-level integration of RL with a dedicated traditional TSC framework (encompassing overflow detection, adaptive overflow phases, and control logic) to resolve overflow-induced gridlock at operational urban intersections, validated by extensive field deployment. We further propose a conceptual technical framework for adapting various RL paradigms such as CoLight series (e.g., extending GNN state embeddings and adding new overflow phases for Advanced-CoLight), FRAP series (e.g., modifying phase competition matrices for MPLight), and Offline RL series (e.g., presetting overflow scenarios in training data for TransformerLight) to our targeted overflow control mechanisms. Compared to emerging large model TSC solutions (e.g., LLMLight~\cite{lai2023llmlight}), OverFlowLight currently offers a more immediately deployable and robust approach, given practical challenges LLMs face in real-time, safety-critical control regarding data requirements, inference speed, hardware demands, and output reliability (e.g., potential for hallucination and inconsistency).

\section{Conclusion}
\label{Sec:Conclu}
This paper addresses vehicle spillback and intersection gridlock during peak hours in real-world scenarios. We propose an overflow-control framework that integrates with both traditional and RL methods. Across 43 intersections in three cities, the framework effectively mitigates spillback through adaptive phase switching while improving efficiency and reducing reliance on manual traffic-police coordination. In future research, we will further explore end-to-end decision optimization and larger-scale coordinated network control.
\clearpage

\normalem
\bibliographystyle{IEEEtran}
\bibliography{submit}

\end{sloppypar}

\clearpage

\appendix
\section{Datasets and Hyperparameters}
\label{appendix_a}
\subsection{Hyperparameters}
The hyperparameters of our method primarily focus on overflow detection as shown in Table~\ref{table: hyperparameter}, which consists of three key parameters: (1) ER (Effective Range), set to 160 meters, representing a relatively close distance to the intersection exit; (2) $\alpha$, the threshold for the stopping time of vehicles across all lanes, set to a moderate value of 3 seconds; and (3) $\beta$, the vehicle queue length within the ER for a single exit lane, optimized to 8 vehicles. These parameters were carefully tuned to balance detection sensitivity and robustness in real-world scenarios.
\begin{table}[h]
\centering

\begin{tabular}{cc} 
\toprule
Hyperparameter & Setting  \\ 
\hline
$\alpha$       & 3s       \\
$\beta$        & 8        \\
$ER$           & 160m     \\
\bottomrule
\end{tabular}
\caption{Hyperparameters}

\label{table: hyperparameter}
\end{table}

\begin{figure}[h]
	\centering
	\begin{minipage}{1\linewidth}	
		\subfigure[Real-world deployment at 18 intersections in City 1]{
			\label{fig:1}
			\includegraphics[width=0.48\linewidth,height=1.2in]{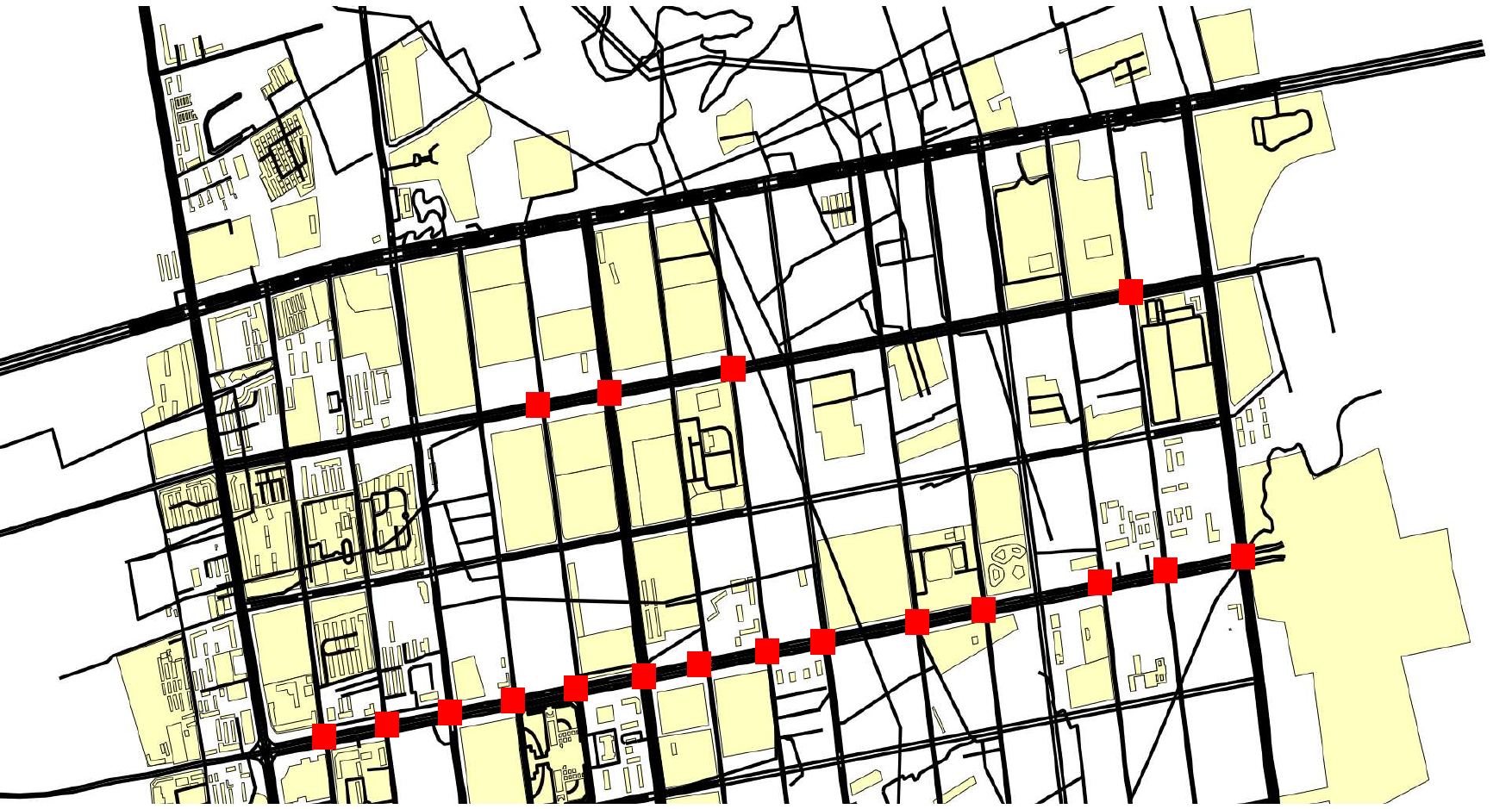}	
		}\noindent
		\subfigure[Real-world deployment at 4 intersections in City 2]{
			\label{fig:2}
			\includegraphics[width=0.48\linewidth,height=1.2in]{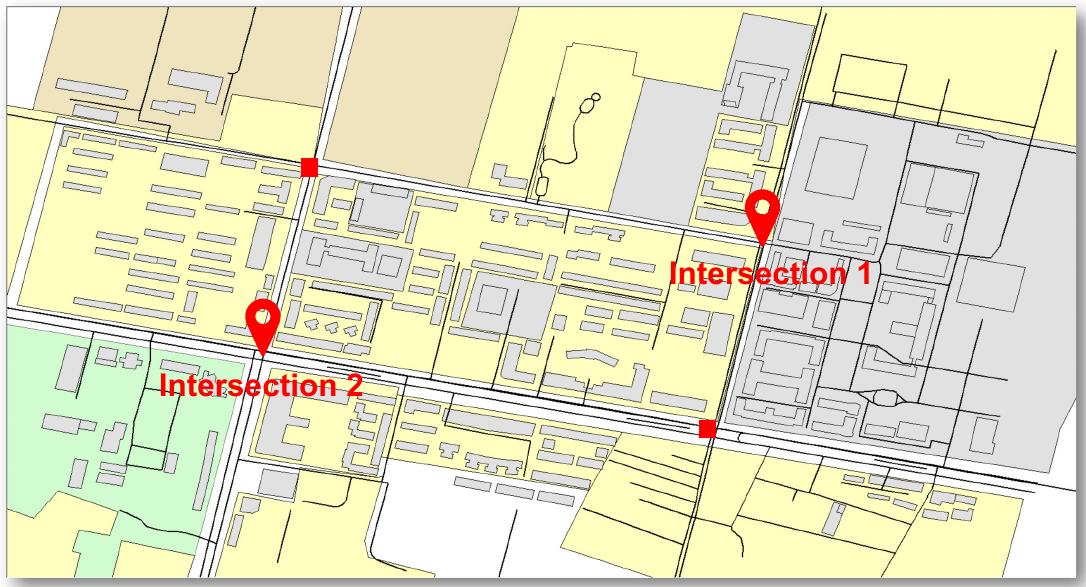}
		}\noindent
        \end{minipage}
        
\begin{minipage}{1\linewidth}	
        \subfigure[Representative Intersection 1 in City 2]{
		\label{fig:2}
			\includegraphics[width=0.48\linewidth,height=1.2in]{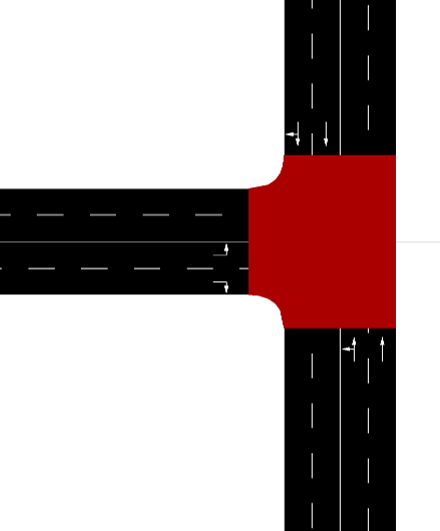}
		}\noindent
         \subfigure[Representative Intersection 2 in City 2]{
			\label{fig:2}
			\includegraphics[width=0.48\linewidth,height=1.2in]{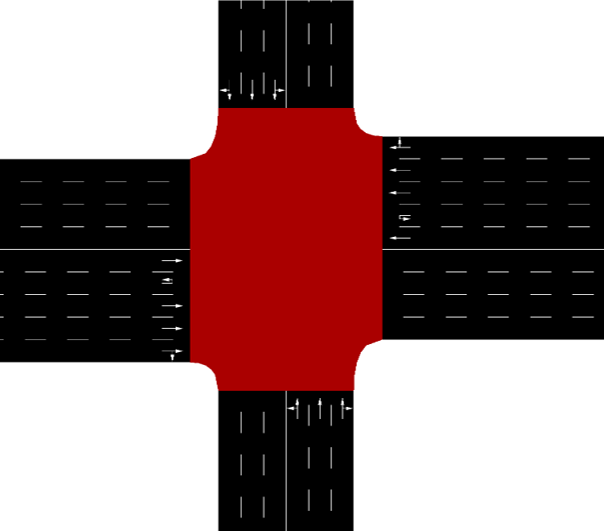}
		}\noindent
	\end{minipage}
	
	\caption{Real-world intersections and road networks.}
\label{fig:roadnet}
\end{figure}

\subsection{Datasets}
We conduct experiments in three real-world cities. The detailed datasets are summarized below.
\begin{itemize}
    \item \textbf{City 1: } The road network has 18 intersections. Each intersection includes either four-way or three-way (T-shaped) configurations, strategically located in the main urban area. Additionally, each intersection exhibits variations in lane lengths and the number of lanes, reflecting the diverse and complex nature of real-world traffic environments.
    \item\textbf{City 2: } The road network has four intersections, all  located within the main urban area of the provincial capital. These intersections are situated in close proximity to a cluster of educational institutions, including universities, colleges, and schools, catering to a student population of nearly 20,000. During peak hours, the substantial increase in vehicular traffic flow and pedestrians frequently leads to overflow phenomena at these intersections, posing significant challenges to urban traffic management and safety.
    \item\textbf{City 3: } The road network has 21 intersections, including both three-way and four-way intersections, all located within the main urban area of the provincial capital.
    \item \textbf{Intersection 1: } It belongs to City 2, is a three-way intersection, with a south entrance, north entrance, and west entrance. The overflow phenomenon is primarily attributed to the insufficient number of traffic lanes, which significantly restricts the road capacity and exacerbates congestion during peak periods.
    \item \textbf{Intersection 2: } It belongs to City 2, is a four-way intersection, with a south entrance, north entrance, east entrance, and west entrance. The overflow primarily results from a mid-block zebra crossing on the northern approach. During school peak hours, pedestrian crossings cause vehicle queues, leading to overflows at the northern entrance due to conflicting pedestrian-vehicle movements.
\end{itemize} 

\subsection{Evaluation Metrics and Settings}
We use the following metrics to evaluate the performance of our proposed methods: 
\begin{itemize}
    \item \textbf{Average Throughput: } The average number of vehicles passing through the intersection during peak overflow periods.
    \item\textbf{Overflow Count: } The number of overflow events during peak overflow periods.
\end{itemize}

Each green signal is followed by a three-second green flash and
a three-second yellow signal time to clear the vehicles at an intersection. For all metrics, we collect data over a three-month period and calculate their average values to ensure statistical reliability and robustness in our analysis.

\section{Additional Experiments}
\label{appendix_b}
\begin{figure*}[t]
	\centering
	\begin{minipage}{1\linewidth}	
		\subfigure[Intersection 1 overflow direction $d_W$ warning]{
			\label{fig:1}
			\includegraphics[width=0.48\linewidth,height=2.4in]{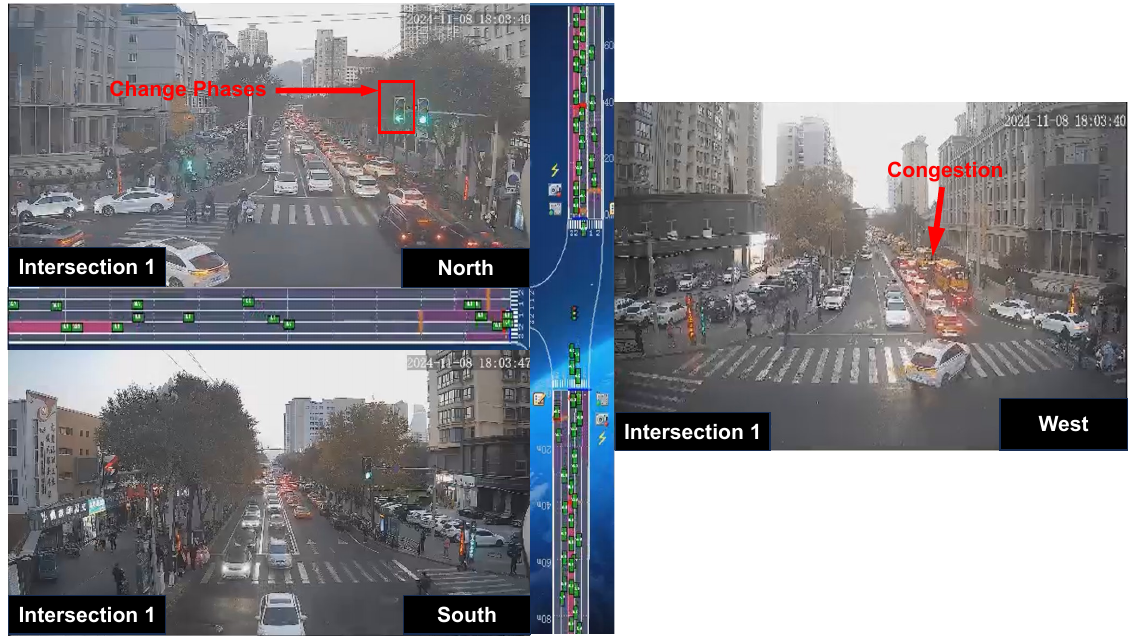}	
		}\noindent
		\subfigure[Intersection 1 change phase to $op_N$ and mitigate overflow]{
			\label{fig:2}
			\includegraphics[width=0.48\linewidth,height=2.4in]{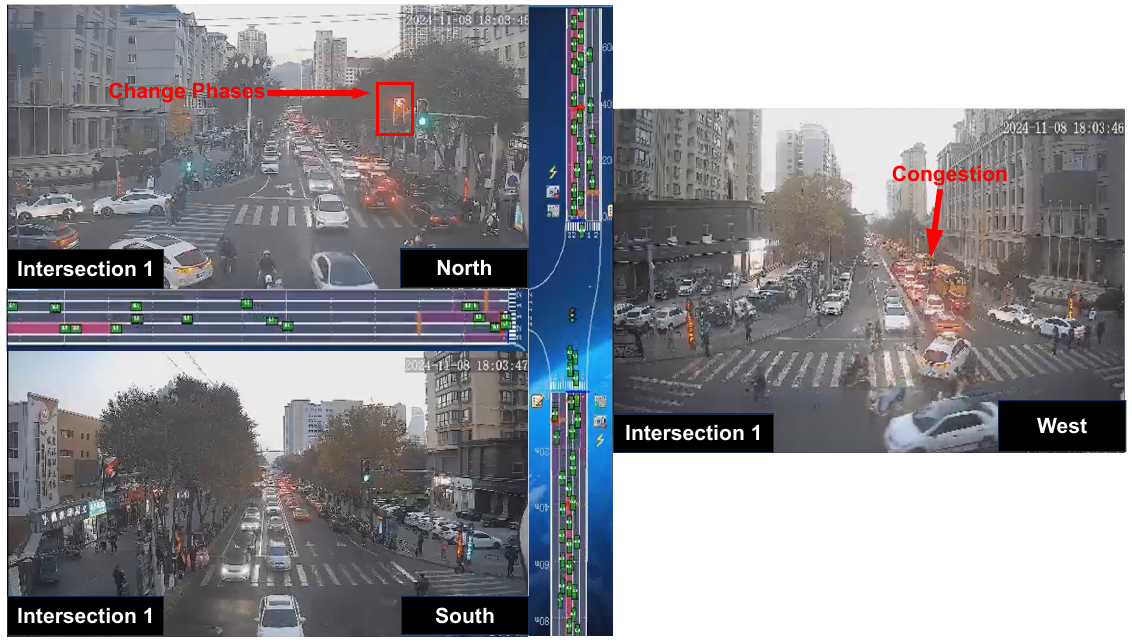}
		}\noindent
	\end{minipage}
	\vskip -0.3cm 
    
        

	\caption{Real-world signal-control snapshots at Intersection 1 integrated with our proposed overflow framework.}
	
	\label{fig:intersection3}
\end{figure*}

This section details further experimental analyses and validations undertaken to assess the OverFlowLight system's capabilities, robustness, and long-term viability. It expands upon the core experimental results by investigating sensor performance under challenging operational conditions and longitudinal operational efficacy.

\subsection{Ablation Study}
\begin{table*}[t]
\centering
\caption{Ablation study: overflow events and frame ratios across representative dates.}
\label{tab:ablation-all}
\begin{tabular}{llrrrrrrrr}
\toprule
 &  & \multicolumn{2}{c}{04-11} & \multicolumn{2}{c}{12-10} & \multicolumn{2}{c}{11-25} & \multicolumn{2}{c}{04-02} \\
\cmidrule(lr){3-4} \cmidrule(lr){5-6} \cmidrule(lr){7-8} \cmidrule(lr){9-10}
Config & Metric & Evt & Rate & Evt & Rate & Evt & Rate & Evt & Rate \\
\midrule
Full OverFlowLight & & 200 & 54.77\% & 195 & 42.68\% & 176 & 36.30\% & 5 & 99.38\% \\
w/o OPM & & 209 & 55.54\% & 202 & 43.19\% & 205 & 36.82\% & 3 & 99.66\% \\
w/o Overflow Detection & & 4 & 6.47\% & 1 & 0.68\% & 1 & 1.46\% & 0 & 0.00\% \\
w/o Overflow Phase Construction & & 355 & 32.43\% & 317 & 25.54\% & 160 & 14.92\% & 51 & 57.08\% \\
\midrule
Avg Waiting & & \multicolumn{2}{c}{16.36} & \multicolumn{2}{c}{12.84} & \multicolumn{2}{c}{7.23} & \multicolumn{2}{c}{37.95} \\
Avg Zero-Speed Vehicles & & \multicolumn{2}{c}{14.38} & \multicolumn{2}{c}{11.60} & \multicolumn{2}{c}{7.65} & \multicolumn{2}{c}{29.94} \\
\bottomrule
\end{tabular}
\end{table*}

\begin{table*}[t]
\centering
\caption{Sensitivity to $\alpha$ with $\beta=5$ fixed.}
\label{tab:alpha-all}
\resizebox{\textwidth}{!}{
\begin{tabular}{r|rrrr|rrrr|rrrr|rrrr}
\toprule
 & \multicolumn{4}{c|}{04-11 (9 windows)} & \multicolumn{4}{c}{12-10 (9 windows)} & \multicolumn{4}{c}{11-25 (4 windows)} & \multicolumn{4}{c}{04-02 (2 windows)} \\
\cmidrule(lr){2-5} \cmidrule(lr){6-9} \cmidrule(lr){10-13} \cmidrule(lr){14-17}
$\alpha$ & Evt & $\sigma$ & Rate & $\sigma$ & Evt & $\sigma$ & Rate & $\sigma$ & Evt & $\sigma$ & Rate & $\sigma$ & Evt & $\sigma$ & Rate & $\sigma$ \\
\midrule
1  & 15.11 & 15.71 & 92.77\% & 14.44 & 15.89 & 12.11 & 93.57\% & 5.66 & 35.50 & 19.05 & 86.22\% & 8.07 & 2.00 & 1.41 & 99.82\% & 0.26 \\
2  & 15.00 & 15.22 & 92.31\% & 14.79 & 14.89 & 10.45 & 92.98\% & 6.01 & 31.00 & 16.39 & 85.62\% & 8.44 & 3.00 & 0.00 & 99.35\% & 0.26 \\
3  & 15.33 & 14.78 & 91.86\% & 15.14 & 15.22 & 10.05 & 92.42\% & 6.32 & 31.00 & 15.73 & 85.09\% & 8.75 & 3.00 & 0.00 & 99.88\% & 0.79 \\
5  & 15.67 & 14.97 & 90.94\% & 15.78 & 15.44 & 10.57 & 91.30\% & 6.94 & 30.25 & 14.31 & 84.05\% & 9.32 & 3.50 & 0.71 & 97.92\% & 1.80 \\
7  & 17.11 & 15.09 & 89.99\% & 16.40 & 16.33 & 10.45 & 90.17\% & 7.55 & 30.00 & 13.71 & 83.04\% & 9.86 & 4.50 & 2.12 & 96.93\% & 4.39 \\
10 & 18.33 & 16.01 & 88.46\% & 17.35 & 17.11 & 10.17 & 88.36\% & 8.50 & 30.00 & 14.02 & 81.56\% & 10.66 & 5.00 & 4.24 & 95.58\% & 6.27 \\
15 & 19.56 & 15.99 & 85.70\% & 19.11 & 19.33 & 10.65 & 85.05\% & 10.21 & 30.25 & 11.18 & 79.08\% & 11.94 & 5.50 & 4.95 & 93.50\% & 9.31 \\
\bottomrule
\end{tabular}
}
\end{table*}

\begin{table*}[t]
\centering
\caption{Sensitivity to $\beta$ with $\alpha=5$ fixed.}
\label{tab:beta-all}
\resizebox{\textwidth}{!}{
\begin{tabular}{r|rrrr|rrrr|rrrr|rrrr}
\toprule
 & \multicolumn{4}{c|}{04-11} & \multicolumn{4}{c}{12-10} & \multicolumn{4}{c}{11-25} & \multicolumn{4}{c}{04-02} \\
\cmidrule(lr){2-5} \cmidrule(lr){6-9} \cmidrule(lr){10-13} \cmidrule(lr){14-17}
$\beta$ & Evt & $\sigma$ & Rate & $\sigma$ & Evt & $\sigma$ & Rate & $\sigma$ & Evt & $\sigma$ & Rate & $\sigma$ & Evt & $\sigma$ & Rate & $\sigma$ \\
\midrule
2  & 6.22  & 10.45 & 97.30\% & 6.84 & 4.67  & 4.77  & 98.70\% & 1.83 & 9.75  & 4.57  & 97.30\% & 1.66 & 1.00 & 0.00 & 99.38\% & 0.71 \\
3  & 15.67 & 14.97 & 90.94\% & 15.78 & 15.44 & 10.57 & 91.30\% & 6.94 & 30.25 & 14.31 & 84.05\% & 9.32 & 3.50 & 0.71 & 97.92\% & 1.80 \\
5  & 15.67 & 14.97 & 90.94\% & 15.78 & 15.44 & 10.57 & 91.30\% & 6.94 & 30.25 & 14.31 & 84.05\% & 9.32 & 3.50 & 0.71 & 97.92\% & 1.80 \\
7  & 15.67 & 14.97 & 90.94\% & 15.78 & 15.44 & 10.57 & 91.30\% & 6.94 & 30.25 & 14.31 & 84.05\% & 9.32 & 3.50 & 0.71 & 97.92\% & 1.80 \\
10 & 15.67 & 14.97 & 90.94\% & 15.78 & 15.44 & 10.57 & 91.30\% & 6.94 & 30.25 & 14.31 & 84.05\% & 9.32 & 3.50 & 0.71 & 97.92\% & 1.80 \\
\bottomrule
\end{tabular}
}
\end{table*}

\noindent\textbf{Key observations.} Disabling the overflow detector reduces detected events to near zero, confirming that the detector does not over-trigger. Removing OPM consistently increases overflow events, showing that safe action masking is necessary under multi-direction simultaneous overflow. Removing overflow-phase construction has the largest effect at high density, where fixed base phases cannot clear blocked exits in time. For threshold sensitivity, $\alpha=3$--$5$ provides the most stable behavior, while results are almost identical for $\beta\ge 3$, indicating that the default $\beta=5$ is robust.

\subsection{Sensor Robustness in Diverse Conditions}
Experimental evaluations of the OverFlowLight system's sensor suite assess its performance and robustness when subjected to a spectrum of challenging environmental and operational conditions.

\textbf{Validation in Adverse Meteorological Conditions.} 
A critical aspect of real-world efficacy is the system's perceptual accuracy during adverse weather. Field tests, leveraging data from 1.5 years of continuous commercial operation, have been conducted to evaluate sensor performance during heavy precipitation (rain, snow), fog, and airborne particulates.


\subsection{Long-term Operational Stability}
The OverFlowLight system has been in continuous commercial deployment for over 1.5 years, providing a substantial period of real-world operation. Data gathered throughout this extended deployment is utilized in this section to empirically validate the system's long-term operational stability, reliability, and sustained performance. This operational tenure has spanned diverse traffic patterns and seasonal variations, including winter snowfall and extended rainy periods, with the system's consistent performance earning official recognition from local transportation authorities.

\section{More Implementation Detail}
\subsection{More RL Method Implementation Detail}
As existing simulators cannot simulate overflow, we showcase how OverflowLight works with RL methods:
\begin{itemize}
    \item \textbf{CoLight Series (CoLight, Advanced-CoLight, CosLight):} Using Graph Neural Networks for state embedding, incorporating exit-lane data (speed, capacity) to detect overflow. Extend the action space to include overflow phases for real-time congestion mitigation.
    \item \textbf{FRAP Series (MPLight):} Expand the phase competition matrix to include overflow phases, prioritizing overflow mitigation when congestion is detected.
    \item \textbf{Offline Series (TransformerLight, DiffLight):} Integrate overflow phases in offline training to handle peak-hour congestion, leveraging pre-deployed TSC algorithms with overflow control for data collection.
\end{itemize}
  
By integrating with OverflowLight, these methods effectively prevent gridlock in high-traffic scenarios where traditional approaches fail.

\subsection{Hardware Implementation} In our experimental environment, the perception and decision algorithm primarily utilizes the Nvidia V100 for model training. 
The perception trained model is then exported as an RKNN format and deployed on a 6 TOP CPU with a tri-core architecture for model inference. The decision-making algorithm is inferred on the CPU consisting of 4 Cortex-A76 and 4 Cortex-A55 cores. The hyperparameters are as Table~\ref{tab:hyperparameters_2}.
\begin{table}[h]
\centering
\caption{Training hyperparameters}
\begin{tabular}{l c}
\hline
\textbf{Parameter} & \textbf{Value} \\
\hline
Learning Rate & 0.001 \\
Batch Size & 64 \\
Discount Factor & 0.99 \\
Epsilon & 0.1 \\
Epsilon Decay & 0.995 \\
Minimum Epsilon & 0.01 \\
Target Network Update Frequency & 10 \\
Epoch & 100 \\
\hline
\end{tabular}
\label{tab:hyperparameters_2}
\end{table}

\subsection{Detailed Deployment Analysis}
To facilitate reproducibility and real-world adoption, we will release the perception and control modules upon acceptance at \url{https://github.com/OpenTraffic-Team}. The tables below summarize the deployment budget and scaling strategy used to translate the system from pilot intersections to city-scale operation.

\begin{table*}[t]
\centering
\caption{Per-intersection deployment cost breakdown (excluding human-operation cost).}
\label{tab:component_costs}
\scalebox{0.8}{
\begin{tabular}{lccccl}
\toprule
Component & Unit Cost & Quantity & Total Cost & Mode & Deployment Insights \\
\midrule
mmWave radar & USD 400--500 & 4 & USD 1,600--2,000 & Radar mode & Robust to lighting and weather; strong night-time and occlusion performance \\
Camera & USD \phantom{0}50--100 & 4 & USD \phantom{0}200--\phantom{0}400 & Both modes & Low cost; can reuse existing infrastructure but is lighting-sensitive \\
Edge device (RK3588) & USD 200 & 1 & USD 200 & Both modes & Lightweight inference; CPU/NPU supports on-site perception and control \\
\midrule
Total & --- & --- & USD 2,000--2,600 & Radar mode & High-reliability deployment used in our current field system \\
Total & --- & --- & USD \phantom{0}400--1,000 & Camera-only mode & Low-cost scalable deployment without radar \\
\bottomrule
\end{tabular}
}
\end{table*}

\begin{table*}[t]
\centering
\caption{Perception and control deployment configurations.}
\label{tab:compute_costs}
\scalebox{0.8}{
\begin{tabular}{lllll}
\toprule
Category & Configuration & Deployment Mode & Hardware Requirement & Cost / Intersection \\
\midrule
Perception + Control & Lightweight vision (4$\times$ YOLO) + RL & Single intersection (edge) & RK3588 CPU/NPU & USD 200 \\
Perception + Control & Lightweight vision (4$\times$ YOLO) + RL & Multi-intersection (shared) & Central server for $\sim$10 intersections & USD 120 (average) \\
Control only & RL + OverFlowLight & Both modes & CPU on edge or server & Included \\
\bottomrule
\end{tabular}
}
\end{table*}

\begin{table*}[t]
\centering
\caption{Estimated city-scale deployment budget under different sensing modes.}
\label{tab:city_budget}
\begin{tabular}{lcccl}
\toprule
City Scale & \# Intersections & Radar Mode & Camera-only Mode & Deployment Insights \\
\midrule
Small city & 10 & USD 20K--26K & USD \phantom{0}4K--10K & Suitable for pilot deployment \\
Medium city & 50 & USD 100K--130K & USD 20K--50K & Balanced cost-performance tradeoff \\
Large city & 200 & USD 400K--520K & USD 80K--200K & Scalable with hybrid deployment \\
\bottomrule
\end{tabular}
\end{table*}

\noindent\textbf{Practical scaling strategy.}
The deployment cost is dominated by sensors rather than control or compute. In a fully decentralized setup, each intersection can be equipped with a low-cost edge device that runs 4-camera YOLO-based perception and RL control locally. In a centralized setup, multiple intersections can share a server; for example, a USD 1,200 server shared across 10 intersections reduces compute cost to roughly USD 120 per site while supporting multi-stream inference. In practice, a hybrid sensing strategy is the most effective: radar is installed at high-risk or congestion-prone intersections, while standard intersections adopt camera-only deployment. This hybrid design achieves an estimated 40\%--60\% cost reduction while maintaining robustness, scalability, and real-time performance.
\subsection{Sim-to-Real Gap}
During peak periods with heavy pedestrian crossings (e.g., school dismissal), mass pedestrian flows dynamically block vehicle movement at exits. CityFlow fails to account for these pedestrian-vehicle interactions. When multiple approach lanes (e.g., 5 lanes) converge into fewer exit lanes (e.g., 2 lanes), the resulting capacity imbalance causes exit saturation. Current simulator queue models cannot properly represent this compression effect. Non-motorized vehicles (bikes/scooters) competing for exit lane space create overflow conditions. SUMO's IDM car-following model, which assumes perfect lane discipline, cannot simulate this realistic interference pattern. These gaps motivate our real-world validation approach.

The sim-to-real problem in TSC is mainly limited by state observation. Real-world noise causes model instability during training. To address this, we use high-precision radar and compressed sensing techniques in the perception module to reduce transmission noise. Additionally, normalization during model training helps minimize noise interference.

In the OverFlowLight module, we use 80GHz millimeter-radar as a cost-effective. By integrating event cameras and visual anonymization, we enhance privacy while capturing key pedestrian and vehicle data. This multi-sensor approach reduces deployment costs to just a few hundred dollars per intersection, enabling large-scale smart traffic system adoption.

\subsection{Generalization}
The proposed framework exhibits strong generalization capability:
\begin{itemize}
    \item \textbf{Intersection structures.}
    The system supports both \textbf{3-way (T-junctions)} and \textbf{4-way intersections}. The two detailed case-study intersections are one 3-way and one 4-way site, and the overall 43-intersection deployment spans all three cities. The OPM scales naturally by remapping detected overflow directions to safe phases, independently of the intersection topology.

    \item \textbf{Lane imbalance.}
    Real-world intersections often exhibit heterogeneous lane distributions (e.g., 1 vs.\ 3 lanes). Since our rules operate on lane-level features such as vehicle count, waiting time, and speed, the framework adapts effectively to lane imbalance without requiring parameter tuning or structural modification.

    \item \textbf{Algorithm compatibility.}
    The OPM is decoupled from the underlying control algorithm. While DQN is used for demonstration, the same mechanism is directly applicable to \textbf{PPO, FRAP, CoLight}, and rule-based controllers. This confirms that OPM serves as a \textbf{general policy-level constraint} rather than an algorithm-specific component.
\end{itemize}

\subsection{DQN Architecture and Training Details}
The DQN module adopts a lightweight feed-forward neural network optimized for real-time traffic control. The overall design is summarized as follows:
\begin{itemize}
    \item \textbf{Input Representation (24 dimensions):}
    Lane-level vehicle count, waiting time, and occupancy signals extracted from radar and camera sensors.

    \item \textbf{Network Architecture:}
    \begin{itemize}
        \item Linear(24, 64) with ReLU activation
        \item Linear(64, 32) with ReLU activation
        \item Linear(32, 2) as the output layer for estimating Q-values of two overflow phase actions
    \end{itemize}

    \item \textbf{Training Dataset:}
    An offline dataset of approximately \textbf{12{,}000 samples} was collected from the deployed FuzzyLight system across diverse traffic scenarios.

    \item \textbf{Optimization Details:}
    The network is trained for 100 epochs using mini-batch updates with the Adam optimizer (learning rate $1\times 10^{-3}$) and a discount factor $\gamma = 0.95$.

    \item \textbf{Shared-Policy Deployment:}
    The current DQN implementation uses one shared policy rather than a separate policy per intersection. This is possible because lane-level features provide a consistent input representation across heterogeneous layouts, while the OPM converts each site's local geometry into a site-specific feasible action subset.

    \item \textbf{Safe Policy Execution via OPM:}
    Action outputs are post-processed by the rule-based OPM mask to discard unsafe phases in real time. The final action is selected as
    \[
        a^{*} = \arg\max_{a \in \mathcal{A}_{\text{safe}}} Q(s, a),
    \]
    where $\mathcal{A}_{\text{safe}}$ is the action subset permitted by the OPM. Inference and masking incur an average overhead of $59.83\,\mathrm{ms}$, leading to an overall decision latency of approximately $106\,\mathrm{ms}$ per step. This design achieves a practical balance between \textbf{safety guarantees} and \textbf{adaptive control}.
\end{itemize}

\end{document}